\documentclass{bmvc2k}
\usepackage{wrapfig}
\usepackage{siunitx}
\usepackage{textcomp}
\usepackage{lipsum, booktabs}
\usepackage{multirow}
\usepackage{tikz}

\title{Learning Spatio-Temporal Features with Two-Stream Deep 3D CNNs for Lipreading}

\addauthor{Xinshuo Weng}{http://www.xinshuoweng.com}{1}
\addauthor{Kris Kitani}{http://www.cs.cmu.edu/~kkitani}{1}

\addinstitution{
Robotics Institute\\
School of Computer Science \\
Carnegie Mellon University\\
Pittsburgh, USA
}

\runninghead{Weng and Kitani}{Two-Stream Deep 3D CNNs for Lipreading}

\def\checkmark{\tikz\fill[scale=0.4](0,.35) -- (.25,0) -- (1,.7) -- (.25,.15) -- cycle;} 

\begin{document}

\maketitle
\begin{abstract}
We focus on the word-level visual lipreading, which requires recognizing the word being spoken, given only the video but not the audio.
State-of-the-art methods explore the use of end-to-end neural networks, including a shallow (up to three layers) 3D convolutional neural network (CNN) + a deep 2D CNN (\emph{e.g.}, ResNet) as the front-end to extract visual features, and a recurrent neural network (\emph{e.g.}, bidirectional LSTM) as the back-end for classification.
In this work, we propose to replace the shallow 3D CNNs + deep 2D CNNs front-end with recent successful deep 3D CNNs --- two-stream (\emph{i.e.}, grayscale video and optical flow streams) I3D.
We evaluate different combinations of front-end and back-end modules with the grayscale video and optical flow inputs on the LRW dataset. 
The experiments show that, compared to the shallow 3D CNNs + deep 2D CNNs front-end, the deep 3D CNNs front-end with pre-training on the large-scale image and video datasets (\emph{e.g.}, ImageNet and Kinetics) can improve the classification accuracy.
Also, we demonstrate that using the optical flow input alone can achieve comparable performance as using the grayscale video as input. Moreover, the two-stream network using both the grayscale video and optical flow inputs can further improve the performance.
Overall, our two-stream I3D front-end with a Bi-LSTM back-end results in an absolute improvement of 5.3\% over the previous art on the LRW dataset.
\end{abstract}

\vspace{-4mm}
\section{Introduction}
\vspace{-2mm}
Word-level visual lipreading, the ability to recognize the word being spoken from visual information alone, is a challenging task for both human and machine due to its ambiguity. As the McGurk effect introduced in \cite{McGurk1976}, different characters can produce similar lip movement (\emph{e.g.}, `p' and `b'). These characters, called homophones, are difficult to be determined from visual cues alone. However, people have demonstrated that this ambiguity can be resolved to some extent using the context of neighboring characters in a word (\emph{i.e.}, word-level lipreading). Therefore, modeling the temporal context between frames is very important to recognize the nearby characters. In this work, we apply recent successful deep 3D convolutional neural networks (CNNs) to learn the temporal context for word-level visual lipreading.

Traditionally, researchers have approached the visual lipreading by a two-stage pipeline, including feature extraction from the mouth region and classification using the sequence model, in which the process in two stages are independent. The most common feature extraction approaches use a dimension reduction or compression method, with the most popular being the Discrete Cosine Transform (DCT) \cite{Xiaopeng2006, Pass2010, Potamianos2001, Potamianos2003}. This yields us compact image-based features of the mouth. In the second stage, a sequence model, such as the Hidden Markov Model (HMM) \cite{Stewart2014, Shao2008, Estellers2012}, is used to model the temporal dependency from the extracted features for classification.

\begin{figure}
\centering
\includegraphics[trim=0cm 4.3cm 3cm 0cm, clip=true, width=0.95\linewidth]{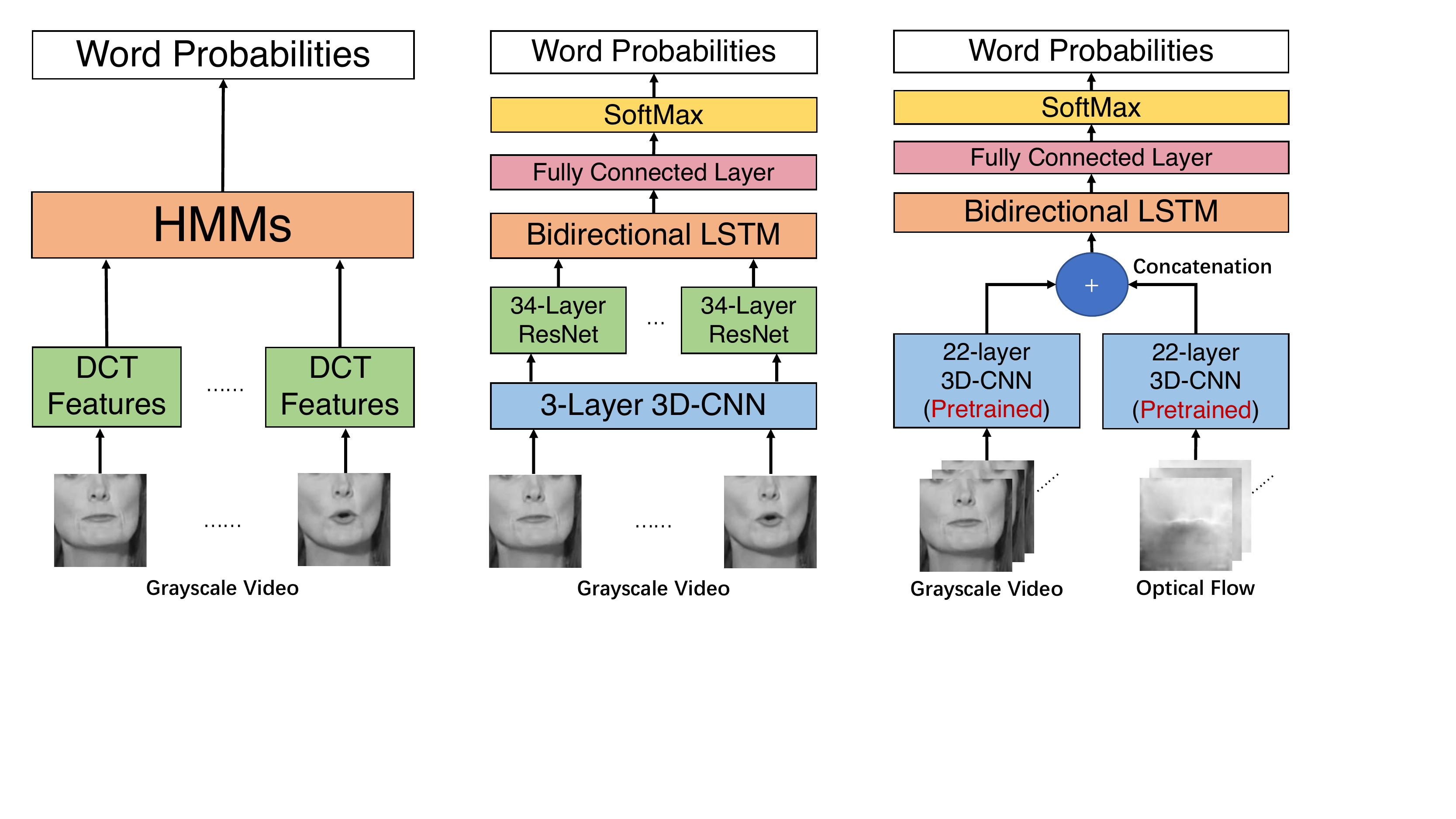}
\vspace{-6mm}
\caption{\textbf{Comparison of the Lipreading Pipeline}: traditional two-stage pipeline (\textbf{left}), end-to-end neural network based pipeline (\textbf{middle}) and the proposed pipeline (\textbf{right}). The proposed one is the first investigating deep 3D CNNs beyond three layers for lipreading.}
\label{fig:teaser}
\end{figure}

Recently, deep learning approaches for visual lipreading \cite{Wand2016, JoonSonChung2016, Stafylakis2017, Petridis2018, Noda2014, Takashima2016, Almajai2016, Ninomiya2015, Sui2015, Petridis2016} have been proposed and achieved the state-of-the-art performance, in which the two-stage pipeline is replaced with the end-to-end trainable neural networks. Similar to the traditional methods, a sequence of the cropped mouth is fed into a front-end to extract the visual features, which are then passed to a back-end to model the temporal dependency for classification. As the gradients computed from the loss can flow back from the back-end module to the front-end module, the entire network is end-to-end trainable such that the learned features are more suitable for lipreading than the standalone features used in the traditional methods. Specifically, existing works usually use the combination of a shallow 3D CNN (up to three 3D layers) and a deep 2D CNN (\emph{e.g.}, ResNet \cite{He2016}) as the front-end, and use the bi-directional recurrent neural network (\emph{e.g.}, Bi-LSTM) as the back-end.

However, it is commonly accepted that image-based features extracted from the 2D CNNs are not directly suitable for video task (\emph{e.g.}, visual lipreading). Intuitively, it is more natural to use the 3D CNNs as the front-end to learn spatial-temporal features. Surprisingly, only a few works \cite{Assael2017, Stafylakis2017, Afouras2018, Petridis2018}, to the best of our knowledge, have explored the use of 3D CNNs for lipreading. Also, only a shallow 3D CNN (up to three 3D convolutional layers) is used in those works, in which the network architecture design is contradictory to our common knowledge that a deep neural network is expected to do better than a shallow one. We conclude that the deep 3D CNNs with a large number of parameters will easily over-fit on lipreading datasets and perform even worse than the model with the shallow 3D CNNs + deep 2D CNNs front-end, which does not have a severe over-fitting problem when training on standard lipreading datasets. In addition, we observe that no prior work has explored the use of optical flow for visual lipreading, although the optical flow has been proved to be very useful in many video tasks \cite{Simonyan2014,bonneel2015blind,janai2017computer}.

In this paper, we present the first word-level lipreading pipeline \emph{with the use of deep 3D CNNs and optical flow}. We compared the proposed pipeline with traditional and recent deep learning based pipelines in Figure \ref{fig:teaser}. To the best of our knowledge, this is the first work investigating the deep 3D CNNs beyond three layers and optical flow input for lipreading. For deep 3D CNNs, we use a 3D version of 22-layer Inception network (I3D \cite{Carreira2017}), first proposed for video action recognition, as our front-end. To handle the over-fitting problem caused by the increased number of parameters from the 3D kernels, we borrow the idea from \cite{Carreira2017} to inflate the pre-trained weights of the 2D Inception network on the ImageNet \cite{krizhevsky2012imagenet} into three dimensions as the initialization of the 3D kernels. Moreover, with the conclusion from \cite{Hara2017} that a 3D CNN can be fine-tuned to achieve better performance on other video datasets when first pre-trained on large-scale video datasets, we thus perform a second-round pre-training on the Kinetics \cite{Carreira2017} dataset, which is a large-scale video dataset. 

We evaluate different combinations of front-end and back-end modules on the challenging LRW dataset. The experiments show that, compared to the shallow 3D CNNs + deep 2D CNNs front-end, the deep 3D CNNs front-end with pre-training on the large-scale image and video datasets (\emph{e.g.}, ImageNet and Kinetics) can improve the classification accuracy for visual lipreading. Also, to show the effectiveness of the optical flow input in lipreading, we replace the conventional grayscale video input with pre-computed optical flow. The conclusion is that using the optical flow input alone can achieve comparable performance as using the grayscale video as input. Furthermore, a two-stream network using both the grayscale and optical flow inputs can further improve the performance. Overall, our two-stream I3D front-end with a Bi-LSTM back-end results in an absolute improvement of 5.3\% over previous state-of-the-art methods on the LRW dataset.

Our contributions are summarized as follows: (1) we present the first work which applies the deep 3D CNNs beyond three layers for word-level visual lipreading; (2) we demonstrate that pre-training on the large-scale image and video dataset is important when training the deep 3D CNNs for lipreading; (3) we show empirically that using the optical flow as an additional input to the grayscale video in a two-stream network can further improve the performance; (4) our proposed two-stream I3D front-end with a Bi-LSTM back-end achieves the state-of-the-art performance on standard lipreading benchmark.

\vspace{-0.4cm}
\section{Related Work}
\vspace{-2mm}

\noindent\textbf{Word-Level Visual Lipreading.}
Existing methods to approach word-level lipreading by using visual information alone can be mostly split into two categories. The first category method, shown in the Figure \ref{fig:teaser} (left), is mainly composed of two separate stages involving the feature extraction from the mouth region and classification using the sequence model, in which the process in two stages are independent. Variants in this category are different in having a different data pre-processing procedure or using different feature extractors and classifiers. \cite{Pass2010} proposes a DCT coefficient selection procedure, which prefers the higher-order vertical components of the DCT coefficients, to achieve better performance when the variety of the hand pose is large in the dataset. \cite{Potamianos2001} proposes a new cascaded feature extraction process including not only the DCT features but also a linear discriminant data projection and a maximum likelihood-based data rotation. \cite{Xiaopeng2006} proposes a PCA-based method to reduce the dimension of the DCT-based features.

Although this category methods with the two-stage pipeline have made significant progress over the decades, all methods in this category separate the feature extraction process from the classifier's training process, causing that the extracted features might not be the best suitable for word classification.

The second category methods, shown in Figure \ref{fig:teaser} (middle), approach the visual lipreading based on the recent advance of deep learning. The major difference from the first category methods is that the two-stage process no longer exists. Instead, the entire system composed of a front-end and back-end neural network is end-to-end trainable, and thus the learned features are more related to the specific task that the network is trained on. \cite{Noda2014} is the first work which proposes to use the CNNs as the front-end to replace the independent feature extractor used in the traditional methods. \cite{Almajai2016} proposes a speaker-adaptive training procedure to achieve a speaker-independent lipreading system with the use of deep neural networks. \cite{Wand2016} is the first proposing to use the LSTM \cite{Hochreiter1997} as the back-end for classification and has shown a significant improvement over the traditional classifiers. \cite{JoonSonChung2016} takes advantage of the proposed large-scale lipreading dataset and train a VGG-like front-end for lipreading. Beyond \cite{JoonSonChung2016}, \cite{Chung2018} also shows the effectiveness of adding the LSTM module at the end for word classification. Furthermore, \cite{Stafylakis2017, Assael2017} propose to use a shallow 3D CNN (up to three 3D convolutional layers) in combination with a strong deep 2D CNN --- ResNet --- as the front-end in the lipreading system. Similarly, \cite{Assael2017} uses three 3D convolutional layers for sentence-level lipreading. Beyond using visual features, \cite{Petridis2018} proposes a neural network to extract the audio features and fuses with the visual features.

Surprisingly, we find that the existing methods in this category have only explored to use the deep 2D CNNs, shallow 3D CNNs or their combination for video feature extraction in lipreading. In contrast, our proposed method is the first work investigating the benefit of using a deep 3D CNN as the front-end for visual lipreading and has demonstrated a significant improvement over previous state-of-the-art methods.

\vspace{2mm}\noindent\textbf{Neural Network Architectures for Video Feature Extraction.} Researchers have developed CNN architectures specifically for video understanding for a long time. To re-use the ImageNet \cite{krizhevsky2012imagenet} pre-trained weights, \cite{Karpathy2014} proposes to extract features independently from each frame using a 2D CNN and then pool the features to predict the action. More satisfying approaches \cite{Donahue2015, Ng2015} add a recurrent neural network as the back-end, such as a LSTM, instead of a pooling layer to capture the temporal dependency and compensate the disadvantages of the 2D CNNs. In addition, as the optical flow can encode the motion information directly, \cite{Simonyan2014, Feichtenhofer2016, Wu2016, Zhu2018} propose the two-stream networks, which has an additional optical flow stream to extract the motion features using the 2D CNNs. On the other hand, a natural extension of the 2D convolution for video data is the 3D convolution. \cite{Tran2015, Ji2013, Taylor2010} propose to use the 3D CNNs for video action recognition. Beyond using the shallow 3D CNNs, 3D-ResNet \cite{Hara2018, Liu2018} explores the 3D version of the deep 2D CNNs --- ResNet --- by replacing all 2D operations with their 3D counterparts, outperforming the 2D CNNs+LSTM and two-stream 2D CNNs paradigms. To alleviate the over-fitting problem when training the deep 3D CNNs, P3D-ResNet \cite{Qiu2017} proposes to decompose the 3D convolution with a 2D convolution along the spatial dimension and 1D convolution along the temporal dimension. As a natural combination of the two-stream networks and deep 3D CNNs, \cite{Carreira2017} proposes a two-stream deep 3D CNN (I3D) pre-trained on the large-scale video dataset --- Kinetics, achieving the state-of-the-art performance on action recognition. Furthermore, \cite{Manglik2019} employs the same I3D network for collision prediction. In this paper, we apply a similar idea (\emph{i.e.}, the two-stream deep 3D CNNs) and aim at transferring the success to visual lipreading.

\vspace{-4mm}
\section{Lipreading Framework}
\vspace{-2mm}

\begin{wrapfigure}{r}{0.35\linewidth}
\centering
\includegraphics[width=0.9\linewidth]{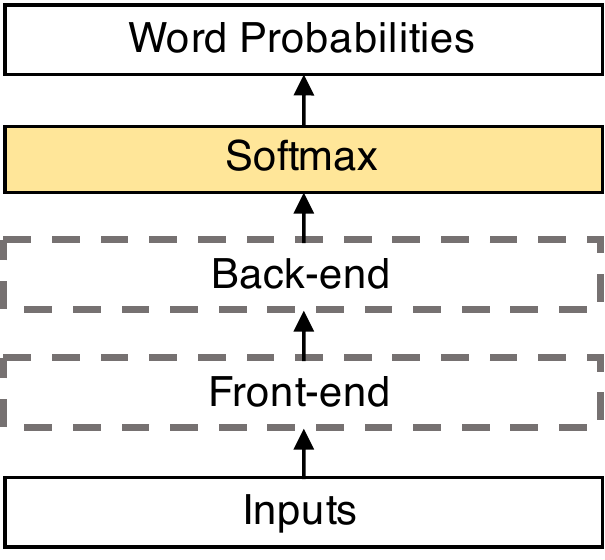}
\vspace{-0.3cm}
\caption{Lipreading framework}
\label{fig:framework}
\vspace{-0.5cm}
\end{wrapfigure}

In Figure \ref{fig:framework}, we decompose the lipreading framework into four parts: (a) inputs, which can be a sequence of grayscale images, optical flow data, or the combination of them; (b) a
front-end module, to extract the features from the inputs; (c) a back-end module, to model 
the temporal dependency and summarize the features into a single vector which represents the raw score for each word; (d) a softmax layer, to compute the probabilities of each word. With above definition of the lipreading framework, we claim that most of the existing lipreading methods, and the proposed method, fall into it.
In this paper, we focus on the inputs and front-end modules which will be introduced in the following sections. We will show how different combinations of them will affect the lipreading performance.

\vspace{-3mm}
\subsection{Front-End Modules}
\begin{figure}
\centering
\begin{minipage}[t]{.49\linewidth}
\centering\includegraphics[width=0.7\linewidth]{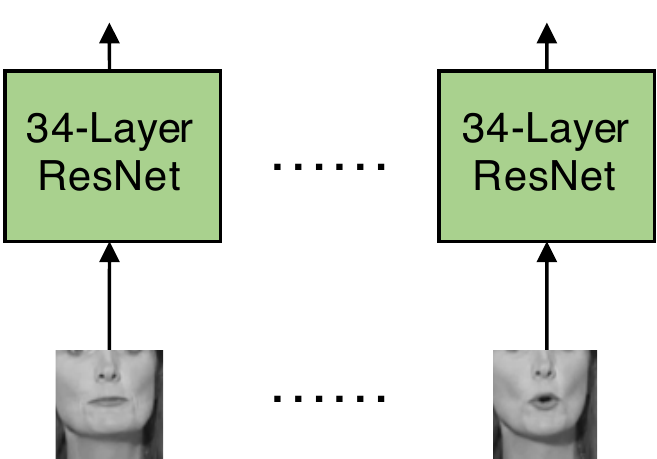}\\
\vspace{-0.7mm}
(a) Deep 2D CNNs
\vspace{1mm}
\end{minipage}
\begin{minipage}[t]{.49\linewidth}
\centering\includegraphics[width=0.59\linewidth]{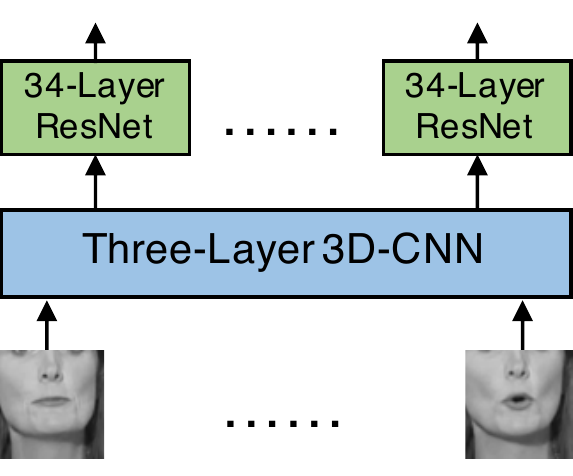}\\
\vspace{-1mm}
(b) Shallow 3D CNNs + Deep 2D CNNs
\vspace{1mm}
\end{minipage}
\begin{minipage}[t]{.49\linewidth}
\centering\includegraphics[trim=0cm 0cm 0cm 0.1cm, clip=true, width=0.65\linewidth]{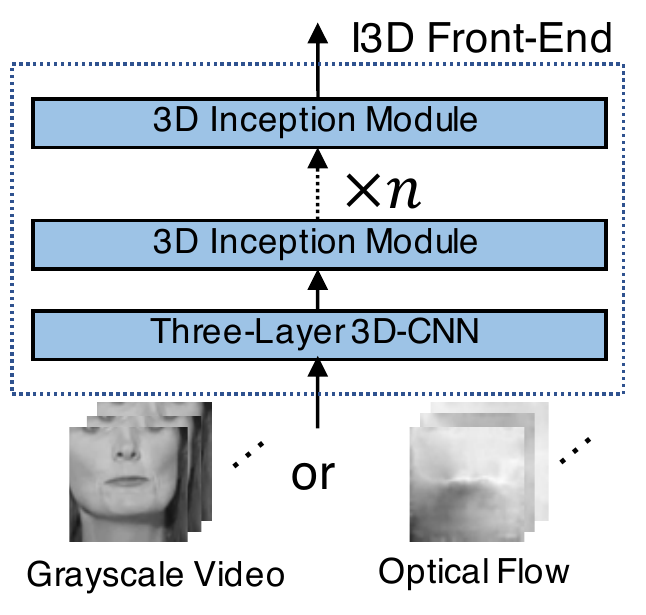}\\
\vspace{-2mm}
(c) Deep 3D CNNs
\end{minipage}
\begin{minipage}[t]{.49\linewidth}
\centering\includegraphics[trim=0cm 0cm 0cm 0cm, clip=true, width=0.82\linewidth]{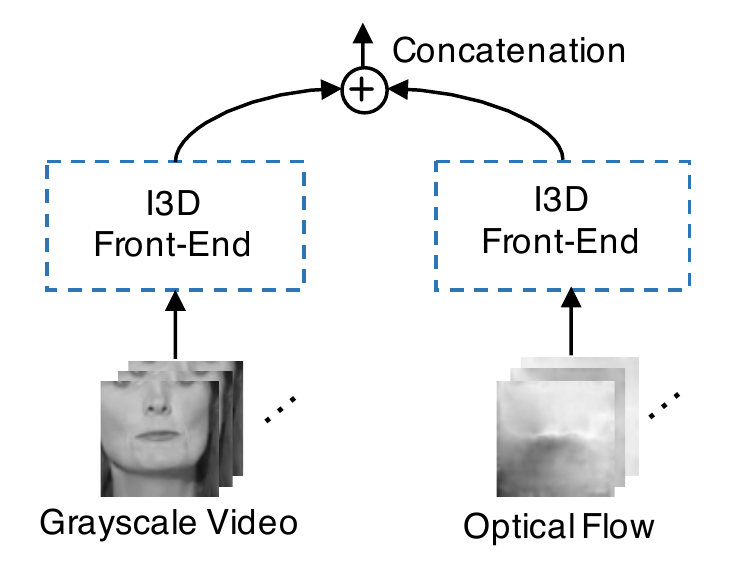}\\
\vspace{-2mm}
(d) Two-Stream Deep 3D CNNs
\end{minipage}
\caption{\textbf{Comparison of the Front-End Modules}: (a) a deep 2D ResNet is used to extract features independently for each frame; (b) a shallow (three-layer) 3D CNN is added before the ResNet; (c) a deep (22-layer) I3D (Inflated 3D ConvNets) front-end learns the spatial-temporal features from either the grayscale video or the optical flow data; (d) a two-stream I3D concatenates the features extracted from both the grayscale video and optical flow data.}
\label{fig:front}
\end{figure}

\vspace{-1mm}
We compare the deep learning based front-end modules for visual lipreading progressively in terms of the complexity of the network. Figure \ref{fig:front} shows the front-end modules that will be discussed in this paper.

\vspace{-3mm}
\subsubsection{Recap: Deep 2D CNNs}
\vspace{-1mm}
The 2D CNNs are originally proposed to solve image-based tasks \cite{Szegedy2015, He2016, krizhevsky2012imagenet, simonyan2014very}. When working with the video data in visual lipreading, we can concatenate the features extracted independently from each frame. Although the features extracted from the deep 2D CNNs front-end only contain the image-based information, the following back-end module such as a recurrent neural network can model the temporal dependency to some extent. We consider the deep 2D CNNs as one of our baseline front-end modules, shown in Figure \ref{fig:front} (a). Specifically, we use the 34-layer ResNet to extract the features from the grayscale video.

\vspace{-3mm}
\subsubsection{Recap: Shallow 3D CNNs + Deep 2D CNNs}
\vspace{-1mm}
In addition to using the deep 2D CNNs, existing works \cite{Stafylakis2017, Assael2017} investigate using a shallow 3D CNN to pre-process the grayscale video before applying the deep 2D CNNs. It is commonly known that the 3D convolution can capture the short-term dynamics and is proven to be advantageous in visual lipreading even when the recurrent networks are deployed for the back-end. However, due to the difficulty of training a huge number of parameters introduced by the three-dimensional kernels, state-of-the-art methods in lipreading have only explored the shallow 3D CNNs with no more than three layers. We consider this shallow 3D CNNs + deep 2D CNNs as another strong baseline front-end module, shown in Figure \ref{fig:front} (b). Specifically, we pass the grayscale video through three 3D convolutional layers with 64, 64 and 96 kernels of a size of 3$\times$5$\times$5, 3$\times$5$\times$5 and 3$\times$3$\times$3 (time/width/height) respectively, each followed by the batch normalization \cite{ioffe2015batch}, Rectified Linear Units (ReLU)\footnote{Each 3D convolutional layer will be followed by the batch normalization and ReLU by default, and will not be mentioned again in the following sections.} and 3D max-pooling layer with a size of 1$\times$2$\times$2. The extracted features are then processed by the 34-layer ResNet on each channel of features.

\vspace{-3mm}
\subsubsection{Deep 3D CNNs}
\vspace{-1mm}

\begin{wrapfigure}{r}{0.38\linewidth}
\vspace{-0.65cm}
\includegraphics[width=\linewidth]{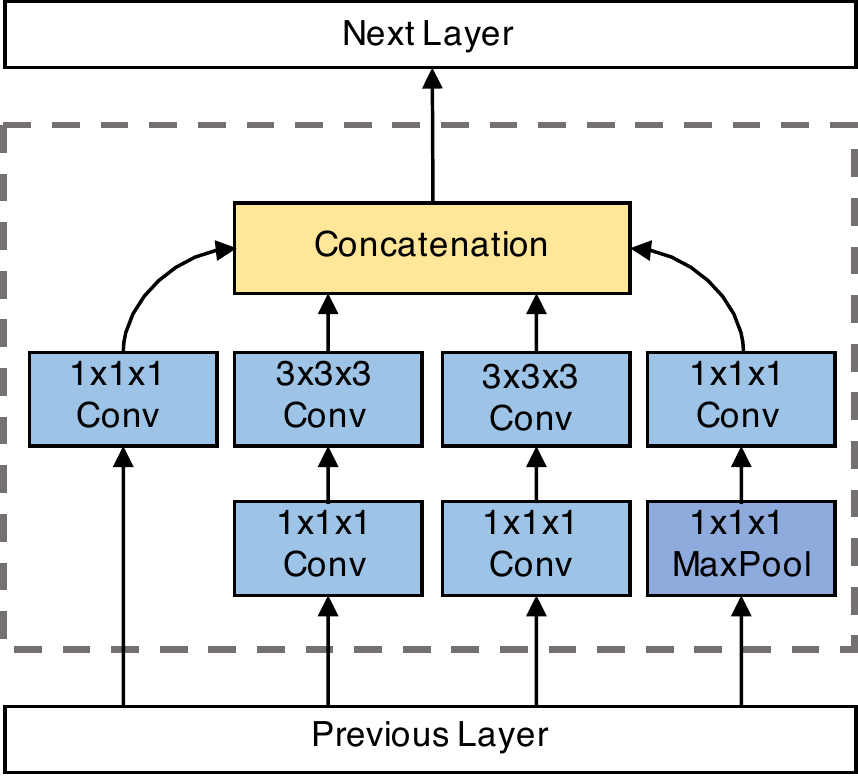}
\vspace{-0.7cm}
\caption{3D inception module.}
\label{fig:3dinception}
\vspace{-0.2cm}
\end{wrapfigure}

Although the shallow 3D CNNs are proven to be helpful, the effect of only using three-layer 3D convolutional layers might not hit the maximum. Also, the temporal dynamics encoded in the features extracted from the shallow 3D CNNs might be impaired by the following 2D CNNs. Therefore, it is natural to use the deep 3D CNNs alone to replace the shallow 3D CNNs + deep 2D CNNs front-end. Specifically, we consider the successful Inflated 3D ConvNets (I3D \cite{Carreira2017}), the state-of-the-art 3D CNNs and first proposed for video action recognition, as our deep 3D CNNs front-end, shown in Figure \ref{fig:front} (c). The I3D front-end contains twenty-two 3D convolutional layers including three 3D convolutional layers with 64, 96 and 192 kernels of a size of 7$\times$7$\times$7, 3$\times$3$\times$3 and 3$\times$3$\times$3 respectively, 3D max-pooling layer with a size of 1$\times$2$\times$2, and a series of 3D inception modules (the 3D extension of the 2D inception \cite{Szegedy2015} module), shown in Figure \ref{fig:3dinception}.

\vspace{-3mm}
\subsubsection{Two-Stream Deep 3D CNNs}
\vspace{-1mm}

\begin{wrapfigure}{r}{0.6\linewidth}
\vspace{-0.5cm}
\centering
\includegraphics[width=0.95\linewidth]{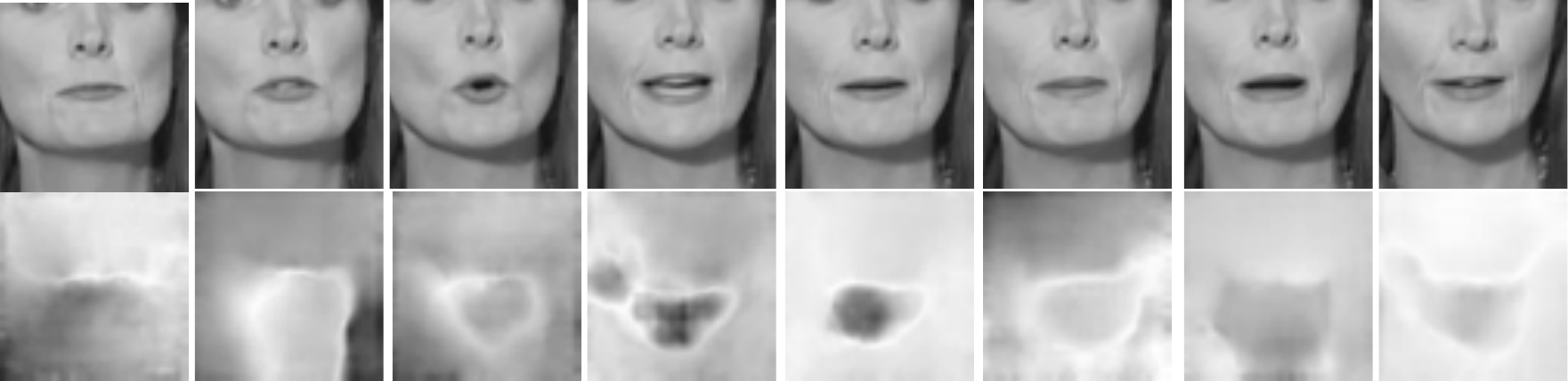}
\vspace{-0.2cm}
\caption{Examples of images and corresponding flow.}
\label{fig:flow}
\vspace{-0.2cm}
\end{wrapfigure}

In theory, deep 3D CNNs are expected to extract strong enough spatial-temporal features from the gray-scale video inputs. In practice, fusing an additional optical flow stream with the grayscale video stream is proven to be useful in many video tasks \cite{Simonyan2014,bonneel2015blind,janai2017computer}, as the optical flow can explicitly capture the motion of pixels in adjacent images, represented by the direction and magnitude of the motion of each pixel. Specifically, we use the PWC-Net \cite{Sun2018} to pre-compute the optical flow. Examples are shown in Figure \ref{fig:flow}. We then use another I3D front-end (employing the same architecture as the I3D front-end in grayscale video stream but not sharing weights) to extract the features from the optical flow and concatenate with the features extracted from the grayscale video stream. We illustrate our two-stream deep 3D CNNs in Figure \ref{fig:front} (d).

\vspace{-3mm}
\subsection{Back-End Modules}
\vspace{-1mm}

\subsubsection{1D Temporal ConvNets (TC)}
\vspace{-1mm}

\begin{wrapfigure}{r}{0.25\linewidth}
\vspace{-7mm}
\centering
\includegraphics[width=0.8\linewidth]{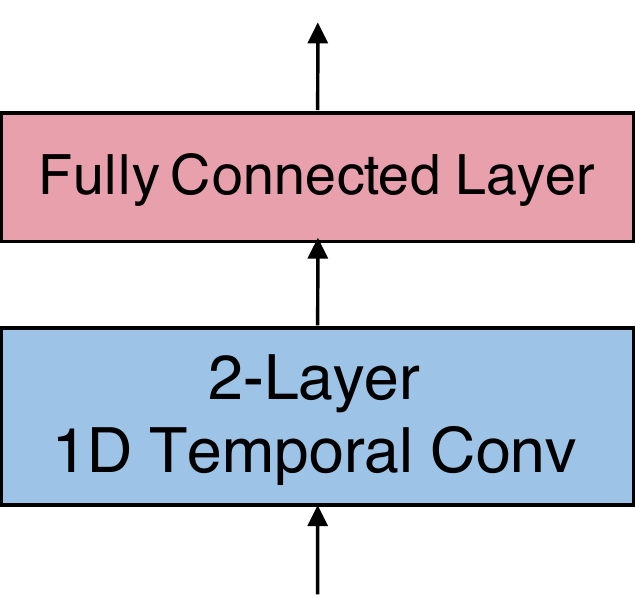}
\vspace{-0.4cm}
\caption{1D Temporal ConvNets back-end.}
\label{fig:back_1d}
\vspace{-0.3cm}
\end{wrapfigure}

In the simplest setting, we can just use the 1D convolution operating along the temporal channel to aggregate the information for word prediction. We consider this 1D temporal ConvNets as our baseline back-end, shown in Figure \ref{fig:back_1d}. Specifically, we use two 1D convolutional layers operating along the temporal dimension followed by a fully connected layer which maps the length of the vector to the number of words in order to compute the raw score for each word. In detail, the first 1D convolutional layer has 512 kernels with a kernel size of 5 and a stride of 2, followed by a max-pooling layer with a kernel size of 2 and a stride of 2. The second 1D convolutional layer has 1024 kernels with the same kernel size and stride as the first one.

\vspace{-3mm}
\subsubsection{Bidirectional LSTM}
\vspace{-1mm}

\begin{wrapfigure}{r}{0.3\linewidth}
\centering
\vspace{-0.65cm}
\includegraphics[width=\linewidth]{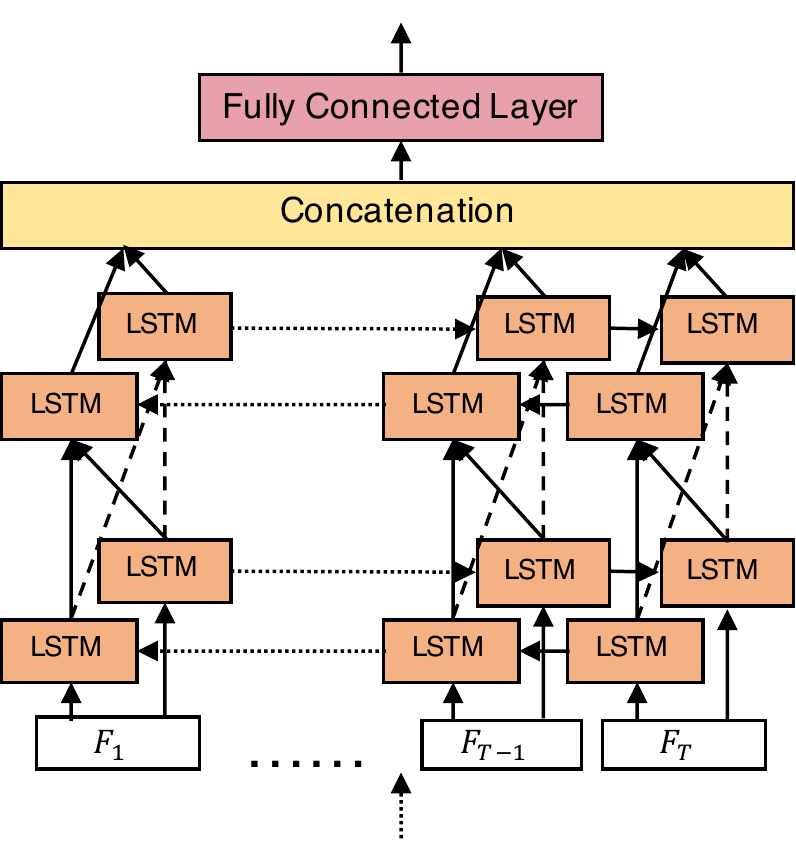}
\vspace{-0.75cm}
\caption{Two-layer bidirectional LSTM back-end.}
\label{fig:back_blstm}
\vspace{-0.2cm}
\end{wrapfigure}

Recurrent neural networks (RNNs) are well known to model temporal dependency and are typical back-end modules used in visual lipreading. 
Among RNNs, LSTM is proven to be useful when dealing with the exploding and vanishing gradient problems.
On the other hand, bidirectional RNNs \cite{schuster1997bidirectional} are the widely used technique allowing RNNs to have both backward and forward information about the sequence at every time step. Combining both, we use the bidirectional LSTM (Bi-LSTM) as our back-end module, shown in Figure \ref{fig:back_blstm}. Specifically, we use a two-layer bidirectional LSTM with a hidden state dimension of 256 for each cell. At the last layer of Bi-LSTM, we concatenate the features from two directions and average the features along the temporal dimension, resulting in the features with a dimension of 512. We then pass the features through a linear layer with a dimension of $512 \times 500$, in order to map to the raw score for each word.

\vspace{-4mm}
\section{Two-Round Pre-Training}
\vspace{-2mm}
ImageNet pre-trained weights are proven to be useful in many image-based tasks. However, it is not possible to directly use the ImageNet pre-trained weights as the initialization of the 3D kernels because the 3D kernels have an additional dimension of weights. Following \cite{Carreira2017}, we inflate the ImageNet pre-trained weights into three dimensions by repeating the weights of the 2D filters N times along the temporal dimension and use inflated weights as the initialization of our I3D front-end. However, such inflated weights pre-trained on the ImageNet can only extract spatial features independently for each channel. With the conclusion from \cite{Hara2017}, when pre-trained on the large-scale video datasets, a deep 3D CNN can eventually achieve better performance on other video datasets after fine-tuning. Therefore, we have a second-round pre-training on the large-scale video dataset --- Kinetics \cite{Carreira2017}, such that the pre-trained 3D kernels can be directly suitable for extracting the spatial-temporal features.

\vspace{-4mm}
\section{Dataset}
\vspace{-2mm}
We train and evaluate the proposed networks on the word-level lipreading dataset LRW \cite{JoonSonChung2016}. The dataset contains short clips of videos (488,766 for training, 25,000 for validation and 25,000 for testing) extracted from the BBC TV broadcasts (\emph{e.g.}, News and Talk Shows). There are 29 frames of grayscale images for each video sequence. LRW dataset is characterized by its high variability regarding speakers and pose. The number of target words is 500, which is an order of magnitude higher than other publicly available databases (\emph{e.g.}, GRID \cite{Czyzewski2017} and CUAVE \cite{Patterson2002}). The other feature is the existence of pairs of words that share most of their visemes. Such examples are nouns in both singular and plural forms (e.g. benefit-benefits, 23 pairs) as well as verbs in both present and past tenses (e.g. allow-allowed, 4 pairs). Perhaps the most difficult part of the dataset is the fact that the target words appear within utterances rather than being isolated. Hence, the network should learn not merely how to discriminate between 500 target words, but also how to ignore the irrelevant parts of the utterance and spot one of the target words. Also, the network should learn how to do so without knowing the word boundaries.

\vspace{-4mm}
\section{Data Pre-Processing and Augmentation}
\vspace{-2mm}
\noindent\textbf{Cropping the Mouth Region}.
As the video provided by the LRW dataset contains the entire faces of the speakers and the surrounding background, which are the irrelevant contexts for
\begin{wrapfigure}{r}{0.5\linewidth}
\vspace{-0.35cm}
\centering
\includegraphics[trim=0.5cm 12cm 9.5cm 0.5cm, clip=true, width=\linewidth]{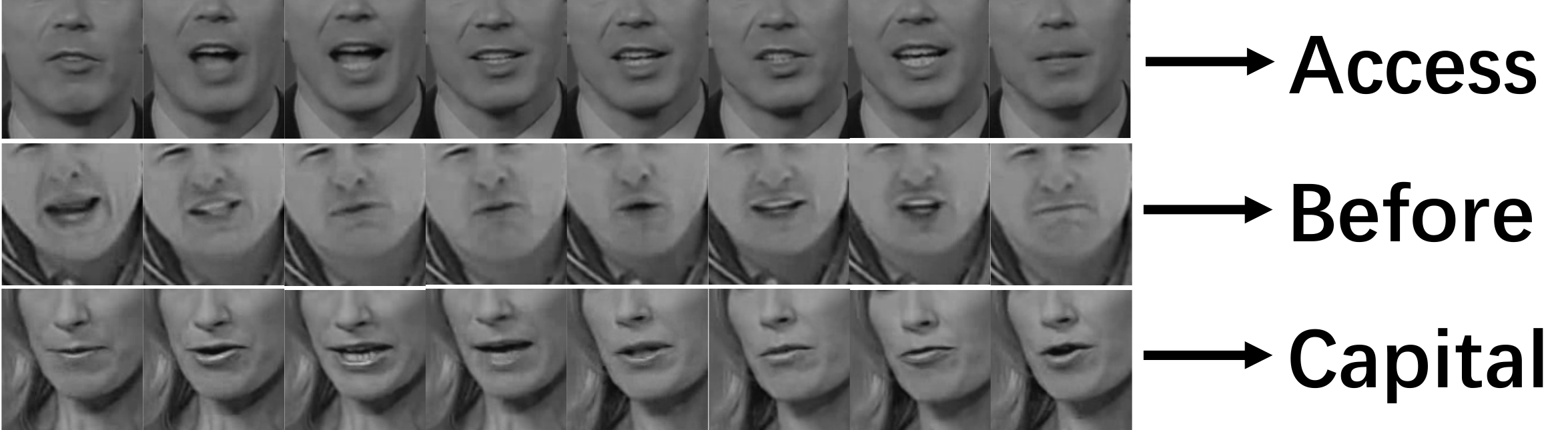}
\vspace{-0.75cm}
\caption{Cropped mouth region and the corresponding labels from the LRW dataset.}
\label{fig:crop}
\vspace{-0.2cm}
\end{wrapfigure}
identifying the spoken words, we crop the original video using a fixed bounding box with the
size of $112 \times 112$ centered at the mouth region, convert the RGB images to grayscale images and normalize them with respect to the overall mean and variance of the dataset. Some random sequences of the cropped mouth from the LRW dataset before the normalization are shown in Figure \ref{fig:crop}.

\vspace{2mm}\noindent\textbf{Data Augmentation.}
We follow the conventional data augmentation mechanisms for visual lipreading during training, including random horizontal flip and cropping (\textpm 5 pixels) so that the training images are not perfectly centered at the mouth which might be the case during the testing. When using the two-stream networks, we augment the data in the same way for both two streams so that the data in two streams are well aligned spatially.

\vspace{-4mm}
\section{Experiments}
\vspace{-2mm}

\subsection{Comparison with State-of-the-Art}
\vspace{-1mm}

\begin{wraptable}{r}{0.39\linewidth}
\vspace{-0.9cm}
\caption{Quantitative comparison of top-1 accuracy on the LRW dataset.}
\vspace{-0.3cm}
\begin{tabular}{ccc}\\\toprule  
Method & Val / \% & Test / \% \\
\hline
\cite{JoonSonChung2016} & - & 61.10\\
\cite{Chung2018} & - & 66.00 \\
\cite{Chung2017} & - & 76.20 \\
\cite{Stafylakis2017} & 78.95 & 78.77\\
\textbf{Ours} & {\bf 84.11} & {\bf 84.07} \\
\bottomrule
\end{tabular}
\label{tab:compare}
\vspace{-0.3cm}
\end{wraptable} 

We summarize the top-1 word accuracy of the state-of-the-art word-level lipreading methods and our proposed one on the LRW dataset in Table \ref{tab:compare}. Our final model, which is composed of a two-stream (grayscale video and flow streams) deep 3D CNNs front-end with two-round pre-training and a bidirectional LSTM back-end, outperforms all previous state-of-the-art methods whose front-ends are the deep 2D CNNs or shallow 3D CNNs with only grayscale video input. On the LRW dataset, our method establishes new state-of-the-art performance and improves the accuracy from 78.77 of prior art \cite{Stafylakis2017}\footnote{We cannot reproduce the accuracy reported in \cite{Stafylakis2017}, which is 83.0 on the test set. One possibility is that we do not perform the three-step training as in \cite{Stafylakis2017} because we want to make a fair comparison with other methods which are trained end-to-end in one step.} to 84.07 on the test set.

\vspace{-3mm}
\subsection{Ablation Study}
\vspace{-1mm}

We conduct the ablative analysis by evaluating the performance of the lipreading networks with different combinations of front-end and back-end modules introduced in previous sections. We use the top-1 word classification accuracy as our evaluation metric on both the validation and testing sets.

\begin{table}
\centering
\vspace{1mm}
\caption{Ablative analysis of top-1 accuracy on the LRW dataset.}
\vspace{1.3mm}
\resizebox{\textwidth}{!}{
\begin{tabular}{c|c|c|c|c|c|c|c|c|c|c}
\toprule
\multirow{2}{*}{Index} & \multicolumn{2}{c|}{Inputs} & \multicolumn{3}{c|}{Front-End} & \multicolumn{2}{c|}{Back-End} & \multirow{2}{*}{Pre-Training} &  \multirow{2}{*}{Val / \%} &  \multirow{2}{*}{Test / \%}  \\
\cline{2-8}
& grayscale & flow & shallow 3D & deep 2D & I3D & 1D TC & Bi-LSTM & & &  \\
\hline
(a) & \checkmark & & \checkmark & \checkmark &  & \checkmark   &  &  & 71.19 & 70.85\\
(b) \cite{Stafylakis2017} & \checkmark & & \checkmark & \checkmark &  & & \checkmark &  &  78.95 & 78.77\\
(c) & \checkmark & & & \checkmark & & & \checkmark & & 75.37 & 75.23 \\
(d) & \checkmark & & & \checkmark & & & \checkmark & ImageNet & 75.69 & 74.98 \\
(e) & \checkmark & & & & \checkmark & & \checkmark & & 59.11 & 59.45 \\
(f) & \checkmark & & & & \checkmark & & \checkmark & Two-Round & 81.73 & 81.52 \\
(g) & & \checkmark & \checkmark & \checkmark & & & \checkmark & & 77.65 & 77.23 \\
(h) & & \checkmark & & & \checkmark & & \checkmark & Two-Round & 82.17 & 82.93 \\
(i) & \checkmark & \checkmark & \checkmark & \checkmark & & & \checkmark & & 80.11 & 80.34 \\
(j) \textbf{Ours} & \checkmark & \checkmark & & & \checkmark & & \checkmark & Two-Round & {\bf 84.11} & {\bf 84.07} \\
\bottomrule
\end{tabular}}
\label{tab:quan}
\vspace{-0.5cm}
\end{table}


\vspace{2mm}\noindent\textbf{Evaluating the Back-End Modules.}
We first focus on the back-end modules by comparing the Bi-LSTM with the 1D temporal ConvNets while fixing the inputs as the grayscale video and the front-end module as the shallow 3D CNNs + Deep 2D CNNs. The results are shown in Table \ref{tab:quan} (a) and (b).
Clearly, using Bi-LSTM as the back-end module is better than using 1D temporal ConvNets. Therefore, we eliminate the 1D temporal ConvNets and only consider the Bi-LSTM as our back-end module in the following experiments.

\vspace{2mm}\noindent\textbf{Evaluating the Effectiveness of Shallow 3D CNNs.}
As the existing works \cite{Stafylakis2017, Assael2017} usually apply a shallow 3D CNN before the deep 2D CNNs in the front-end, we conduct an ablation analysis by removing the shallow 3D CNNs in order to show how important the shallow 3D CNNs are in the front-end. The results are shown in Table \ref{tab:quan} (b)
and (c). Indeed, adding a few 3D convolutional layers in the front-end can have an obvious accuracy improvement from $75.23\%$ to $78.77\%$ on the test set. Also, to make sure the deep 2D CNNs front-end in (c) is well-performed, we run another ablation analysis shown in Table \ref{tab:quan} (d)
\footnote{Here we only do the ablation study on using the ImageNet pre-trained weights to initialize the deep 2D CNNs in (c) but not in (b) because it is commonly accepted that using the pre-trained weights to initialize the intermediate layers usually does not help when there are random initialized lower-level layers (\emph{i.e.}, the shallow 3D CNNs in (b)).}
by using the ImageNet pre-trained weights to initialize the deep 2D CNNs. We find that the results of (c) and (d) are fairly comparable, meaning that the LRW dataset is large enough to train a 34-layer ResNet from scratch without the need for pre-training.

\vspace{2mm}\noindent\textbf{Evaluating the Deep 3D CNNs Front-End and Two-Round Pre-Training.}
As the major contribution of this paper, here we show the advantages of the deep 3D CNNs front-end over the shallow 3D CNNs + deep 2D CNNs front-end. The results are shown in Table \ref{tab:quan} (b) and (f). We show that when the shallow 3D CNNs + deep 2D CNNs front-end in (b) is replaced with the pre-trained I3D front-end in (f), the performance is improved from $78.77\%$ to $81.52\%$ on the test set, meaning that the deep 3D CNNs front-end with the two-round pre-training is more powerful than the shallow 3D CNNs + deep 2D CNNs front-end used in previous state-of-the-art methods on the LRW dataset. On the other hand, network in Table \ref{tab:quan} (e) using the I3D front-end without pre-training (\emph{i.e.}, training from scratch on the LRW dataset) performs even worse than the previous art shown in (b), which shows that training a deep 3D CNN is much harder and can easily over-fit the LRW dataset, and thus the two-round pre-training is important. In the following experiments, we will use the two-round pre-trained weights to initialize the I3D front-end by default.

\vspace{2mm}\noindent\textbf{Evaluating the Optical Flow Input and Two-Stream Network.}
To demonstrate if the optical flow input is effective, we replace the grayscale input in (b) and (f) with the pre-computed optical flow. The results are shown in Table \ref{tab:quan} (g) and (h). We found that both (g) to (b) and (h) to (f) have comparable performance, meaning that either the optical flow or the grayscale video alone can predict the spoken word.
More importantly, we show how the two-stream configuration helps in Table \ref{tab:quan} (i) and (j). When comparing the two-stream network with its single-stream counterpart (\emph{i.e.}, either the grayscale stream or the optical flow stream), the performance is clearly improved no matter the front-end is the shallow 3D CNNs + deep 2D CNNs or the pre-trained I3D. This confirms that the two-stream network is beneficial for lipreading. Overall, our two-stream pre-trained I3D + Bi-LSTM (j) achieves an absolute improvement of 5.3\% over the previous art in (b).

\vspace{-4mm}
\section{Conclusion}
\vspace{-2mm}
We present the first word-level lipreading pipeline using the deep 3D CNNs (beyond three layers) and optical flow. We evaluate different combinations of front-end and back-end modules on the LRW dataset. We show that, with the two-round pre-training, deep 3D CNNs front-end can outperform the shallow 3D CNNs + deep 2D CNNs front-end used in previous works. On the other hand, we found either the optical flow or grayscale video is effective to be used as input. Moreover, the two-stream network using both optical flow and grayscale video as inputs can further improve the performance. Overall, our proposed word-level lipreading method, which is composed of a two-stream (grayscale video and optical flow streams) pre-trained I3D front-end and a Bidirectional-LSTM back-end, achieves an absolute improvement of 5.3\% over the previous art on the LRW dataset.

\vspace{-4mm}
\bibliography{main}
\end{document}